# Personalized Emotion Detection using IoT and Machine Learning

Fiona Victoria Stanley Jothiraj[1][0000-0002-9855-1219] and Afra Mashhadi[1][0000-0003-4631-4438]

[1] University of Washington Bothell, USA
`fiona123@uw.edu`

**Abstract.** The Medical Internet of Things, a recent technological advancement in medicine, is incredibly helpful in providing real-time monitoring of health metrics. This paper presents a non-invasive IoT system that tracks patients' emotions, especially those with an autism spectrum disorder. With a few affordable sensors and cloud computing services, the individual's heart rates are monitored and analyzed to study the effects of change in sweat and heartbeats per minute for different emotions. Under normal resting conditions of the individual, the proposed system could detect the right emotion using machine learning algorithms with a performance of up to 92% accuracy. The result of the proposed approach is comparable with the state-of-the-art solutions in medical IoT.

## 1 Introduction

People interact with each other not only through language but by only using emotions [12]. When words or actions do not seem to convey the message, emotions can facilitate interaction. Psychologists have confirmed six basic emotions that humans could feel or exhibit: happy, sad, anger, fear, surprise, and disgust. Besides this, any other emotion that we notice in everyday lives is a combined feeling of the primary emotions. For example, the emotion of shame comes from the blending of fear and disgust. Emotions are psychological states due to the neurophysiological changes in human beings. The arousal valence model [8] describes the way emotions are distributed in a two-dimensional circular space based on arousal and valence. Valence refers to a positive or negative feeling and arousal refers to an exciting or calming feeling.

In the recent decade, there has been an increase in 'emotion' research ranging from history, psychology to computer science. Specifically, the mobile computing community has focused on detecting emotions from variety of sensors that monitor body as well as relying on qualitative data collected through active experience sampling methods such as notifications [12]. Machine Learning approaches have focused on variety of techniques including those that incorporate video stimuli and video games [1,6,2]. However, majority of the past research concentrates on ordinary people and exclude those with Alexithymia condition.

Alexithymia is a medical condition, and a person diagnosed with this condition often feels emotionally detached. They find it hard to identify, describe and communicate emotions with other people. Autism Spectrum Disorder (ASD) or autism is a complex disability, usually occurring during childhood which inhibits the person's ability to communicate with emotion and creates difficulties with social interaction.
Individuals who have autism spectrum disorder and other complex disabilities face difficulties in their day-to-day lives interacting with other people. They rely on gestures, pictures or drawings, emotive sounds, and other non-verbal communication techniques. A system tailored to help such people with special needs would be a win in the space of technology and medicine. There seems to be a gap in technology to help such individuals be part of a social gathering, and existing solutions rely on natural language processing.

The problem being solved by this project is to understand emotions without verbal cues or facial expressions and create a system to react to them accordingly. The project revolves around an IoT system that consists of biomedical sensors, minicomputer, and cloud computing service that handle and manipulate the sensor data. Developing an IoT system that helps such individuals would be of great cause since it helps such patients and the circle of people surrounding their lives. By harnessing the power of technology, autistic patients can feel comfortable if their emotions are clearly understood and converted into useful information, such as creating the right ambiance in terms of music and lighting.

## 2      Related Work

There has been significant research going on in the field of medical internet of things since it is at the intersection of computer science and medicine. The technique of obtaining useful information from biomedical sensors was derived from the work of Mickael and Paul in [1]. The proposed model in their research relies on detecting and distinguishing the universal emotional expressions which are joy, sadness, anger, fear, disgust, and surprise. Their experiment using low-cost biomedical sensors which capture live data from participants of the experiment who are under video stimuli. The video clips tried to simulate both the visual and the auditory sensory modalities of the 35 participants. The preliminary results from a support vector machine (SVM) algorithm achieve an accuracy of 89%.

Research done by Zhai, Jing & Barreto and Armando [2] approach the problem slightly different and make use of other machine learning classifiers like Naïve Bayes and Decision Trees. Their work used signals to detect stress in the experimental subjects. Results concluded that there was indeed a strong correlation between the physiological signals monitored and the emotional state when subject to a stress stimulus.

The Emotion Recognition System [5] uses a pulse sensor, temperature and ECG sensor that collects data, uses a microcontroller (Arduino) and sends the processed data to the cloud. The work then uses ML algorithms such as KNN and Naïve Bayes classifier to detect the emotion of the individual.

Similarly, there have been numerous works indicating that analysis of physiological signals is a possible approach for emotion recognition. These include signals derived from the heart (ECG or EKG), muscles (EMG), brain (EEG), skin and eye movement (EOG). From these signals, it is possible to model and classify emotions due to the uniqueness of each emotion in terms of arousal and valence [11]. Table 1 shows the approaches and results of different research works.

**Table 1.** Experiments and results of related work

| Research | Sensors | Experiment | Algorithm | Accuracy (%) |
|---|---|---|---|---|
| [1] | Heart rate, Skin conductance | Video stimuli-based sensor data | SVM | 89.58 and 85.57 |
| [6] | Electrodermal, ECG, Photoplethysmography | Video + game-based stimuli | Naïve Bayes | 53.9 |
| [2] | GSR, Blood volume pulse, Pupil diameter, Skin temperature | Computer game | Naïve Bayes | 78.65 |
| | | | Decision Tree | 88.02 |
| | | | SVM | 90.10 |
| [5] | Heart rate, Temperature, Skin conductance, Muscle tension, Brain waves (ECG) | - | Naïve Bayes | 84.70 |
| | | | KNN | 83 |

In our experimental setup, we aim to identify three basic emotions – happy, sad, and angry and understand the physiological activity of each emotion. The key differentiating factor of our work is the use of non-invasive sensors to create a personalized emotion detection system using cloud services.

## 3 System Design

This project aims to provide a complete end-to-end IoT system that parents of such individuals could use to understand the current mood state – happy, sad, or angry. Real-time music/lighting recommendations can be made based on the detected output.
The research assumes that the individual is in a rested condition. The entire system uses cloud services to detect and send personalized recommendations.

The main modules of the IoT system are sensors, microprocessors, and an IoT backend as shown in Figure 1. All the communication between the components happens via MQTT. MQTT is the ideal communication protocol for this solution since it is a lightweight, publish-subscribe network protocol that transports messages from IoT sensors to the cloud.

### 3.1 System Model

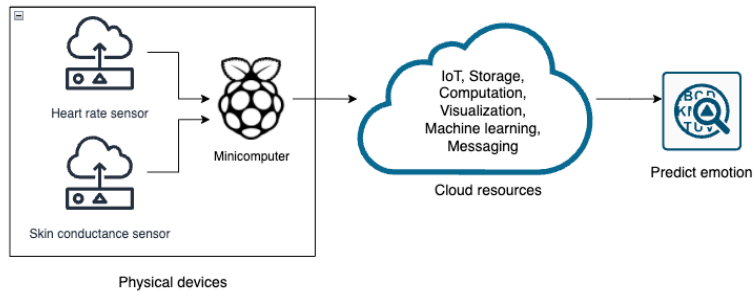

**Fig. 1.** Overview of the IoT system design

**Physical Devices.** The sensors used for the project include a Galvanic skin response sensor (GSR) and heart rate sensor.

A GSR sensor is a biomedical sensor that helps measure sweat gland activity. The GSR sensor outputs the conductive voltage of the skin from which the person's emotion can be determined. The conductive voltage varies based on the amount of sweat produced by the sweat gland controlled by the nervous system—the greater the sweat gland activity, the more perspiration, and less skin resistance. When an individual is calm, not a lot of sweat is secreted, and hence the resistance of the skin is high, whereas the conductive voltage (siemens) is low. More sweat is secreted when individuals are subject to high emotional activity like anger or extreme sadness. Hence, the skin's resistance is low, whereas the conductive voltage is high. The electrodes of the sensors are usually attached to the fingertips, wrist, or foot for accurate readings.

The Heart Rate sensor is used to monitor the heart beats per minute. The sensor consists of an LED in contact with the individual's fingertip. The LED emits light that falls on the veins, monitoring blood flow. From the flow of blood, a person's heartbeats can be calculated. The blood's minor change in reflected light is analyzed to determine the specific heartbeat.

The Raspberry Pi is a low-cost minicomputer that is capable of small computations. Biomedical sensors can be plugged to a Raspberry Pi, which communicates with the cloud computing services using MQTT.

**IoT Cloud Services.** The IoT backend for the project relies on using cloud computing services. Cloud computing is an ideal solution for tasks that generate a large amount of sensor data. When large data is generated, storing, analyzing, and modeling data is best done in the cloud. Some of the benefits of cloud platforms for IoT include scalability, data mobility, security, and cost-effectiveness. The backend provides the necessary resources for data collection, storage, manipulation, machine learning training, messaging, and many other functionalities.

### 3.2 Hardware Requirements

- The Raspberry Pi 3 Model B device has built-in WiFi and Bluetooth connectivity. The specifications of the device are as follows: Quad-Core 1.2 GHz 1 GB RAM.
- Micro SD Card - Raspberry Pi Recommended 32GB MicroSD Card with NOOBS Software.
- Grove Base Hat for Raspberry Pi - Digital/Analog/I2C/PWM/UART port that can be connected to the sensors
- Galvanic skin response sensor with specifications as shown in table 2.
- Ear-clip heart rate sensor with specifications as shown in table 3.

**Table 2.** Skin conductance sensor specification

| Parameter | Value (or) Description |
|---|---|
| Operating voltage | 3.3V/5V |
| Sensitivity | Adjustable via a potentiometer |
| Input Signal | Resistance |
| Output Signal | Voltage, analog reading |
| Finger contact material | Nickel |

**Table 3.** Heart rate sensor specification

| Item | Min | Typical | Max | Unit |
|---|---|---|---|---|
| Voltage | 3.0 | 5.0 | 5.25 | V |
| Work Current | 6.5 | | | mA |
| Length of ear clip wire | 120 | | | cm |
| Measures Range | $\geq 30$/min | | | - |

There is no ADC (Analog-to-Digital converter) in the Raspberry Pi directly. With the help of the built-in MCU STM32, the Grove base can work as a 12-bit ADC. The analog sensor inputs the analog voltage into the 12-bit ADC. The ADC converts the analog to digital data, sent as a digital input to Raspberry Pi through the I2C interface. The Grove base sits on top of the GPIO pins of the Raspberry pi board. The physical sensors are connected to the Analog port of the Grove base hat. Out of the four available analog ports, the galvanic skin response sensor and heart rate sensor are connected to A0 and A4 ports, respectively.

### 3.3 Software Requirements

When the Raspberry Pi is booted with the New Out Of Box (NOOBS) Software for the first time, Raspian OS needs to be installed. The project uses Python3 as the programming language with multiple python libraries for data manipulation, graphs, and model training.

### 3.4 Data

Meaningful data can be extracted from the heart rate sensor and GSR sensor in terms of conductance of the skin and beats per minute of the heart (BPM) of the individual. The participant's fingers are connected to non-invasive sensors to collect raw sensor data. To collect labeled sensor data, the participant is subject to specific video stimuli that evoke happy, sad, or angry emotions. Anomalous data usually collected at the start of the video stimuli are ignored.

Since the range of conductance of skin and BPM varies for everyone, the collected sensor data for each participant is stored separately for machine learning training. This helps provide a personalized emotion detection that is more accurate than if all the data were treated as one. The properties of the conducted experiment and the data collected are shown in table 4.

**Table 4.** Properties of Data Collection

| Item | Description |
| --- | --- |
| Sensors | Heart Rate Sensor, GSR Sensor |
| Experiment | Video Stimuli |
| Length of experiment | 30 seconds – 5 minutes |
| Rate of data collection | 1 record/second |
| Total instances | 11,000 records of sensor data (3,500 instances each for happy, sad, and angry emotion) |

### 3.5 Cloud Services

The IoT backend relies solely on AWS Cloud services. The services used for the project include AWS IoT Core, Amazon S3, Lambda, Kinesis Firehose, Glue, Sagemaker, Cloudwatch, and Simple Notification Service.

### 3.6 Implementation

Python scripts are written to read the raw sensor digital output from these pins and make sense of the collected data. The output of the galvanic skin response sensor can be directly used as resistance. However, the Heart rate requires domain knowledge [3] of the heart rate graphs to understand how the raw sensor data can be converted to BPM (Beats per minute). A sample sensor data record is shown in Figure 4.

To make sure the emotion of patients can be detected, the machine learning algorithm requires thousands of sensor data to analyze and learn the pattern behavior. An experiment is set up for collecting data where the individual is subject to video stimuli that induces happy, sad, or angry feelings. The sensor data is recorded when the individual looks at these videos, and the data is tagged with the respective mood. When the python script is triggered, the individual places his fingertip inside the heart rate sensor and wears the two-finger gloves of the galvanic skin response sensor. The data is collected at a rate of 1 record/ second.

```
{
  "GSR": 2718.2130584192437,
  "BPM": 111.96254489481785,
  "Mood": "Angry",
  "Date": "12/10/2021",
  "Time": "23:23:20"
}
```

**Fig. 2.** json file that shows a sample sensor data record

The project is implemented in two stages: Training and Inference. The flow of data and the services used are entirely different for the two stages. The first stage follows a particular path in collecting sensor data, sending data to the cloud, analysis, and machine learning training as shown in Figure 3. The second stage involves a GUI that triggers a program to collect and send the raw sensor data to the cloud. The cloud workflow then triggers an event that runs the inference model. The predicted result is then sent to the GUI and e-mail as shown in Figure 6.

## 3.7 Training

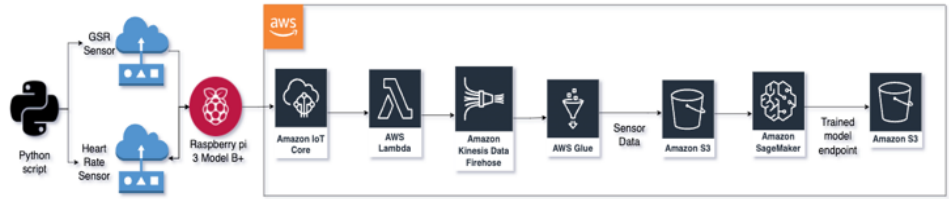

**Fig. 3.** Architecture of the training phase

**AWS IoT Core and AWS Lambda.** In the AWS IoT core service, a 'Thing' named Raspberry_pi is created. A 'Thing' simply refers to a physical device, in this case a Raspberry Pi. To authenticate the Raspberry Pi, device certificates are downloaded and kept securely in the device. These certificates help establish communication between the cloud service and the Raspberry Pi. Like AWS IoT Core 'Thing', another feature named Policies is created to let the AWS IoT service know the names of topics that the device can connect, subscribe, or publish. The device can then publish, subscribe, and receive all topics that start with the "iotsensors" prefix. Rules are used within the AWS IoT core service to let the IoT core know what actions happen when the rule is triggered. The rule is a simple query statement that selects all the raw sensor data sent by the sensors, given by [SELECT * FROM 'iotsensors/train']. When the above conditions are satisfied, actions are taken to send the message to a Lambda Function. The Lambda function named "rounding_op" helps to mathematically round the raw sensor data with up to 3 points of precision.

**AWS Kinesis Data Firehose, AWS Glue.** The lambda function outputs the rounded sensor data comprising skin conductance, heartbeats per minute, timestamp, and mood to the Kinesis Data Firehose service. Kinesis Data Firehose is an extract, transform, and load (ETL) service that captures, transforms, and delivers streaming data to the required services. Using the AWS Firehose, large amounts of real-time data can be streamed without data loss to the required destinations like S3. The project uses a JSON data format in AWS IoT Core which is larger in size and harder to handle during the machine learning training stage. Hence, another service named 'AWS Glue' is used with Kinesis.

In AWS Glue, a crawler is created by pointing it to the individual JSON file containing the sensor data. The crawler crawls through the file, obtains the field names and data types, and stores these details in a glue Table. Using the table information, Kinesis can be configured to perform "conversion of record format" to Apache Parquet file before storing the data onto Amazon S3.

**Amazon S3.** Amazon Simple Storage Service is a cloud object storage service capable of handling and storing large chunks of sensor data. The data outputted from Kinesis Firehose in parquet format is stored in Amazon S3 with the corresponding time stamps. This data serves as the input data for the machine learning model training.

**Amazon Sagemaker.** Amazon Sagemaker is a cloud machine learning platform that helps researchers and developers to train and deploy machine learning models to the cloud. Sagemaker allows Jupyter notebook instances in CPU/GPU mode for this purpose. The data from multiple parquet files on S3 is initially loaded in a Jupyter notebook, and the data is aggregated into one object. The sensor data is cleaned, visualized (figure 4 and figure 5) and Synthetic Minority Oversampling techniques are applied to take care of imbalanced dataset, if any. The data is then separated into vectors and labels for both the training and test set before passing the data through a linear learner algorithm. The vectors, in this case, are the sweat rate (in conductance) and heartbeats per minute, whereas the labels are the tagged emotion for a particular instance.

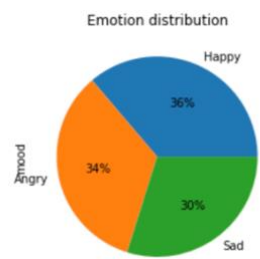

**Fig. 4.** Distribution of dataset for classes – Angry, Happy and Sad emotion

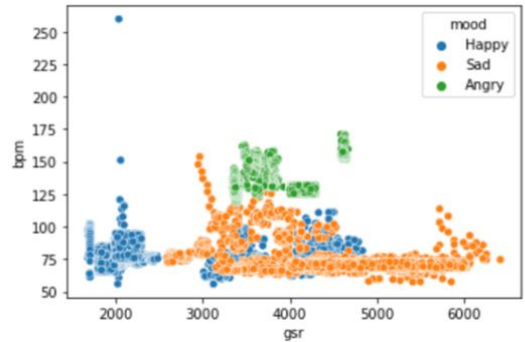

**Fig. 5.** Scatter plot of dataset for varied skin conductance and heartbeat levels

The machine learning algorithm used for the application is a Linear Learner Algorithm of Amazon Sagemaker. Linear models are supervised learning algorithms that best fit a multiclass classification problem. The algorithm fits linear functions that best describe the data and map the inputs to the corresponding output. The model is trained and tested with multiple discrete objectives like F-1 Score, precision, recall, and the confusion matrix. These performance metrics evaluate how well the machine learning model performs against new data. The choice of hyperparameters could significantly impact the performance of a model.

Hyperparameters are parameters whose value is modified to control the learning process. These parameters include loss function, learning rate, number of epochs, and regularization factor.

The notebook instance creates an estimator from the given hyperparameters, fits the training data, and returns a deployed predictor. The deployed predictor is accessible to other AWS cloud services through the Sagemaker endpoint associated with it. The trained model and the logs created during the training process are automatically stored in Amazon S3 for future analysis.

## 3.8 Inference

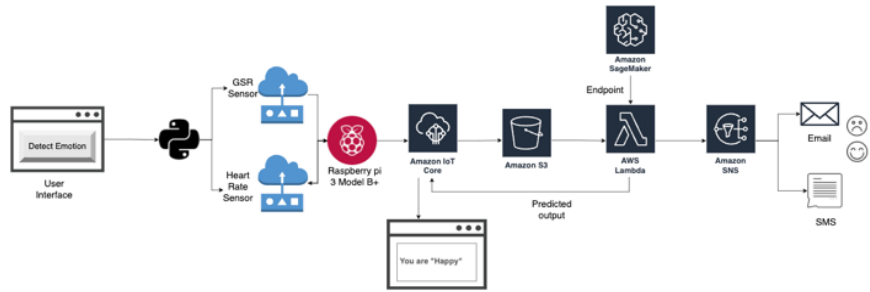

**Fig. 6.** Architecture of the inference phase

**User Interface.** Flask, a micro web framework, is used along with HTML Pages to create a user interface. The UI uses Python, HTML (Hyper Text Markup Language), and CSS (Cascading Style Sheets). The website hosted on the local host consists of a button that triggers the prediction process.

The python script that takes care of the inference stage is triggered when the 'Detect your emotion' button is pressed. The galvanic skin response sensor and the heart rate sensor start collecting the biomedical data from the individual. These sensor values are not tagged with emotion labels since the data acts as the test data for prediction. After a preset duration, all the data is aggregated, and the mean value is computed. The mean value of sweat rate and heart rate data is sent to AWS Cloud after data collection.

**AWS IoT Core.** The mean values are published to the AWS IoT core on a particular topic named 'iotsensors/infer.' Here, rules are set to collect all the received sensor data given by [SELECT * FROM 'iotsensors/infer']. Corresponding actions are triggered that store the input message into an Amazon S3 bucket.

**AWS Lambda.** The Lambda function for the inference stage is configured with an input trigger. Here, the S3 bucket with the collected mean data acts as the trigger. The lambda function starts executing when new data is added to the S3 bucket. The Lambda function reads the sensor data and uses this test data to invoke the Sagemaker endpoint created previously during the training phase. Environment variable named 'Sagemaker_endpoint' is created, which points to the respective linear learner Sagemaker endpoint present under Inference☉Ⅱ Endpoints location of Amazon Sagemaker service. The endpoint returns the predicted emotion with a numeric value as a response. The numeric values are mapped back to output labels as { 0: "Angry", 1: "Happy", 2: "Sad"}.

Using this result and personalized recommendations, output destinations can be configured. This research project uses three outputs to display the prediction emotion – website, SMS, and e-mail. For publishing the result to the SMS and e- mail, another service named Amazon Simple Notification Service is used. However, AWS IoT core is used to publish the result back to the website.

**Output destinations.** The Amazon Simple Notification Service is a fully managed messaging service that can fan out messages to systems like Amazon SQS, Lambda, HTTPS, Email, Text Message as shown in Figure 7. In this service, a topic is created, and email and phone numbers are created as subscriptions. An origination number must also be created under this service for the phone numbers to work.

Similarly, for displaying the result back on the website, the result is published back to AWS IoT Core under the topic 'iotsensors/infer/result'. The inference script that runs on the Raspberry pi device subscribes to the topic and waits for the expected result. The script uses Flasks' render template when a message is received to display it, as shown in Figure 8.

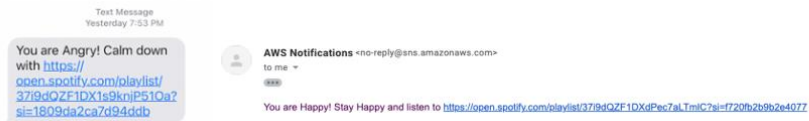

**Fig. 7.** SMS and email notification from AWS SNS service

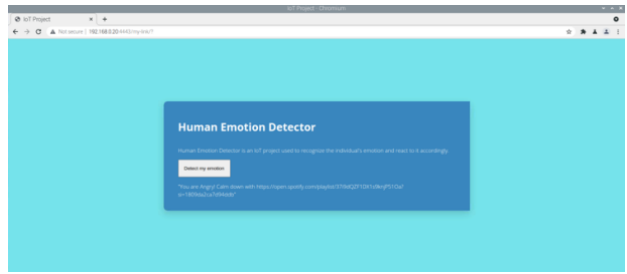

**Fig. 8.** UI Website displaying the output prediction

## 4 Evaluation

The experiment to set up the IoT system involves the collection of large volume of sensor data from the galvanic skin response sensor and heart rate sensor. The experiment participant is subject to video stimuli [10] that induces a particular emotion – happy, sad, or angry. The participant's sensor data is collected under normal resting conditions. The length of each video clip ranges from 30 seconds – 5 minutes. The sensor data is collected for each emotion at a one record/second rate. The machine learning training process is initiated when sufficient sensor data of about 11,000 instances are collected.

To make sure that the IoT system is robust, experiments are evaluated with metrics. The success of a machine learning model can be measured by metrics like accuracy, precision, recall, and F-1 Score. Machine learning libraries like Sagemaker and Scikit-learn are used to ease the process of visualization and training. This system makes use of a multiclass classification problem having balanced classes. A dataset is said to be balanced if the number of instances for each label is almost the same. There are approximately 3500 instances each for angry, sad, and happy output labels in the biomedical sensor data. For a balanced dataset, the best way of evaluating the model is using an Accuracy, ROC curve and confusion matrix.

### 4.1 Accuracy

Model accuracy is the measure used to determine which model is best at identifying patterns and relationships between variables in a dataset based on the input, hyperparameters and training. When a model can generalize better to new 'unseen' data, its predictions are better, resulting in a highly accurate model. The accuracy of the trained linear learner model is 92.5%.

### 4.2 Confusion Matrix

The sensor data consisting of approximately 2200 instances are separated from the training data. This data known as the "test data" is tested against the best machine learning model trained. The confusion matrix shown in Figure 9 and Table 4 gives a comparison between the actual and predicted values. Confusion matrix can be used to measure other useful metrics like AUC-ROC curve that is proportional to performance of the model.

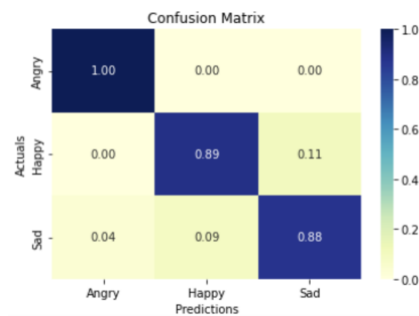

**Fig. 9.** Confusion matrix that shows the distribution of predicted vs. actual values

**Table 4.** Confusion Matrix obtained from results of multiclass classification

| Actual | Prediction | | |
|--------|-------|-------|-----|
|        | Angry | Happy | Sad |
| Angry  | 760   | 0     | 0   |
| Happy  | 2     | 713   | 84  |
| Sad    | 24    | 58    | 587 |

### 4.3 Receiver Operating Characteristic curve

Receiver operating characteristic curve is a graph that shows the performance of a classification model for all threshold values. When the area under the curve is close to 1.0, the trained model is said to be an optimal one. Figure 10 shows the ROC curve obtained after testing the model against 2200 instances.

Another way of evaluating the efficiency of the IoT system is calculating the time taken to predict the output emotion from the time the sensor collects the biomedical data. The mean elapsed time between sensor data reading and receiving the output is 3500 milliseconds. The fast response time is highly beneficial for real-time monitoring of heartbeat and sweat rate that can be used in other medical applications. Since healthcare data is highly sensitive, the future work of this research should focus on adding security layers as per the healthcare privacy regulations.

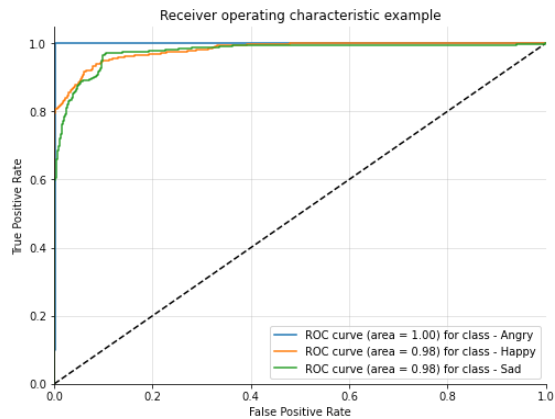

**Fig. 10.** Receiver operating characteristic curve for output classes

## 5     Conclusion and Future work

The research focuses on creating a non-invasive and low- cost IoT system that uses various biomedical sensors to predict the emotion of the individual. It makes use of different emerging technologies like Machine learning, Internet of Things and Cloud computing services. A Raspberry pi 3 along with the heart rate sensor and galvanic skin sensor act as the local system that are connected to the AWS Cloud ecosystem. Various Amazon web services for internet of things, functions, logging, access, streaming, machine learning and messaging are used. The evaluation for the system focused on machine learning model metrics like Accuracy, Confusion matrix and the ROC curve. The system also shows a fast response time of 3 seconds. The potential applications for the medical IoT system are endless, ranging from alerts when stress levels are too high [9], parents aiding autistic children by understanding their emotion [7] and assisting patients with less mobility.

The future work of this project would focus on improving the quality of sensors, integrating more biomedical sensors, prediction of emotions and prediction when the person is not in ideal resting conditions. Although the current sensors serve the purpose, its accuracy cannot be compared to the hospital equipped sensors. Sensors like Skin temperature sensor, Muscle tension tensor, Electrocardiogram (ECG) sensor can be integrated to the existing system that can improve the accuracy of prediction. Emotion recognition is useful in developing human machine interaction systems in the future, which has good potential in the education field [4]. Factors like security, machine learning algorithm accuracy, more emotions, more biomedical sensors, and high volume of data can yield an accurate and efficient solution.